\pdfoutput=1

\documentclass[11pt,a4paper]{article}

\usepackage[]{emnlp2022}

\usepackage{times}

\usepackage{latexsym}
\usepackage{tabularx}
\usepackage{arydshln}
\usepackage{xcolor}

\usepackage[T1]{fontenc}

\usepackage[utf8]{inputenc}

\usepackage{microtype}
\usepackage[normalem]{ulem}
\usepackage{booktabs}
\usepackage{threeparttable}
\usepackage{color,soul}
\usepackage{graphicx} 
\usepackage{tikz}
\usepackage{pgfplots}
\definecolor{DarkBlue}{RGB}{28,81,140}
\definecolor{DarkRed}{RGB}{151,11,28}
\definecolor{DarkGreen}{RGB}{49,147,70}
\definecolor{LightGreen}{RGB}{129,234,108}
\newcommand{\model}{\texttt{RedApt}\xspace}
\makeatletter
  \newcommand\greenuline{\bgroup\markoverwith{\textcolor{green}{\rule[-0.7ex]{2pt}{0.4pt}}}\ULon}
  \UL@protected\def\redwave{\leavevmode \bgroup 
    \ifdim \ULdepth=\maxdimen \ULdepth 3.5\p@
    \else \advance\ULdepth2\p@ 
    \fi \markoverwith{\lower\ULdepth\hbox{\textcolor{red}{\sixly \char58}}}\ULon}
\makeatother

\usepackage{times}
\usepackage{multicol}
\usepackage{multirow}
\usepackage{graphicx}
\usepackage{booktabs}
\usepackage{microtype}
\usepackage{algorithm}
\usepackage{algorithmic}
\usepackage{amsbsy}
\usepackage{soulpos}
\usepackage{amsmath}
\usepackage{multirow}
\usepackage{makecell}

\usepackage{amsmath,soul}
\usepackage{comment}
\usepackage{soulpos}
\usepackage{xcolor}
\usepackage[english]{babel}
\usepackage{blindtext}
\usepackage[shortcuts]{extdash}
\usepackage{amssymb}
\usepackage{amsmath}
\usepackage{hyperref}
\newcommand{\stitle}[1]{\vspace{0.3ex} \noindent{\bf #1}}
\ulposdef{\ulnumaux}{%
   $\underset{\saveulnum}{\rule[-.7ex]{\ulwidth}{.4pt}}$}

\title{\model: An Adaptor for \textsc{wav2vec 2} Encoding \\ Faster and Smaller Speech Translation without Quality Compromise}

\author{Jinming Zhao\ \ \ \ \ \ \ \ \ Hao Yang\ \ \ \ \ \ \ \ \ Gholamreza Haffari\ \ \ \ \ \ \ \ \ Ehsan Shareghi \\ \ \ \
Department of Data Science \& AI, Monash University \\
\texttt{{first.last}@\{monash.edu\}}}

\begin{document}
\maketitle

\begin{abstract}
Pre-trained speech Transformers in speech translation (ST) have facilitated state-of-the-art (SotA) results; yet, using such encoders is computationally expensive. To improve this, we present a novel Reducer Adaptor block, \model, that could be seamlessly integrated within any Transformer-based speech encoding architecture. Integrating the pretrained \textsc{wav2vec 2} speech encoder with \model brings 41\% speedup, 33\% memory reduction with 24\% fewer FLOPs at inference. To our positive surprise, our ST model with \model outperforms the SotA architecture by an average of 0.68 BLEU score on 8 language pairs from Must-C.
\end{abstract}

\section{Introduction}
\label{sec:introduction}

Leveraging pre-trained speech Transformers, such as \textsc{wav2vec 2 (w2v2)}~\cite{baevski2020wav2vec}, in speech-to-text translation (ST) systems have established state-of-the-art~(SotA) results across several languages~\cite{li2020multilingual,ye2021end}. Meanwhile, the high computational cost of such encoders is well-documented~\cite{wu2022performance} and mainly attributed to the self-attention mechanism inside Transformers~\cite{NIPS2017_7181}.

However, speech modality introduces unique challenges compared to text: representation of raw speech signals are orders of magnitude longer\footnote{Speech features, e.g. Mel-Spectrogram, are lower dimensional but result in worse ST quality~\cite{ye2021end}.}, while empirical findings suggest that high-quality ST systems require much deeper encoders compared to text-to-text translation~\cite{wang2020fairseq}. Consequently, training and inference with such encoders is  expensive, in terms of memory and time, even for reasonably powerful hardware.\footnote{Training batch size for a modern ST system~\cite{gallego2021end} could not exceed 1 on a V100 16GB GPU.}

As a mitigation, \citet{li2020multilingual} 
augmented \textsc{w2v2} at the output by CNN layers to reduce the representation length, and \citet{DBLP:conf/interspeech/ZhaoYHS22} proposed a Transformer-based adaptor to shrink a sequence. Yet, the complexity of encoding remains high. \citet{wu2022performance} proposed lower feature dimensions in \textsc{w2v2} which improved efficiency at the expense of performance drop. Earlier studies focused on designing feature selection modules at the input level by using phone labels for merging adjacent vectors~\cite{salesky2020phone}, using dynamic sparsification mechanism~\cite{zhang2020adaptive}, or injecting Connectionist Temporal Classification \cite{gaido2021ctc} to regulate feature passing between layers. These approaches improve speed at inference but are limited as they rely either on hand-crafted features, transcripts, or external modules.\footnote{Pruning~\citep{lai2021parp}, quantization and knowledge distillation~\cite{peng2021shrinking, chang2022distilhubert} approaches are promising directions but they have yet to be applied to ST.}  

In this work we focus on \textsc{w2v2}, as one of the most widely used pre-trained speech encoder for ST, and propose a novel block, Reducer Adaptor~(\model), to reduce the computational load of processing speech sequences through \textsc{w2v2} while improving translation quality. Our approach does not require any additional pretraining or information beyond the audio input, and works similar to adaptor blocks~\cite{HoulsbyGJMLGAG19} placed on top of any layers of a pretrained transformer but trained along with the underlying Transformer on ST task.

Through extensive experiments on 8 language  pairs from Must-C, we show that integrating \model into \textsc{wav2vec}
yields 41\% speedup, 33\% memory savings, and 24\% fewer FLOPs at inference time. 
Meanwhile, our ST model outperforms  existing SotA by 0.68 BLEU score.
To the best of our knowledge, we are the first to target the efficiency of pretrained speech encoders for ST. We hope this to facilitate further improvement across a broader range of speech processing tasks that require pretrained speech encoders.

\section{Reducer Adaptor (\model)}
Our proposed \model has some key properties: (i) can be integrated seamlessly with pretrained speech Transformer such as \textsc{wav2vec 2 (w2v2)}, (ii) flexibly reduces the computational load of encoding (both in terms of memory and time), and (iii) allows to retain the downstream ST performance. While beneficial for  training, \model is in particular useful in the repetitive nature of inference phase. In this section, we will present \model architecture, and show how it can be integrated into \textsc{wav2vec} and a full speech-to-text translation (ST) system.\footnote{Our code is available at \url{https://github.com/mingzi151/w2v2-st}}

\subsection{Architecture of \model}\label{redapt_arch}
\begin{figure}[t]
  \centering
    \includegraphics[scale=0.375]{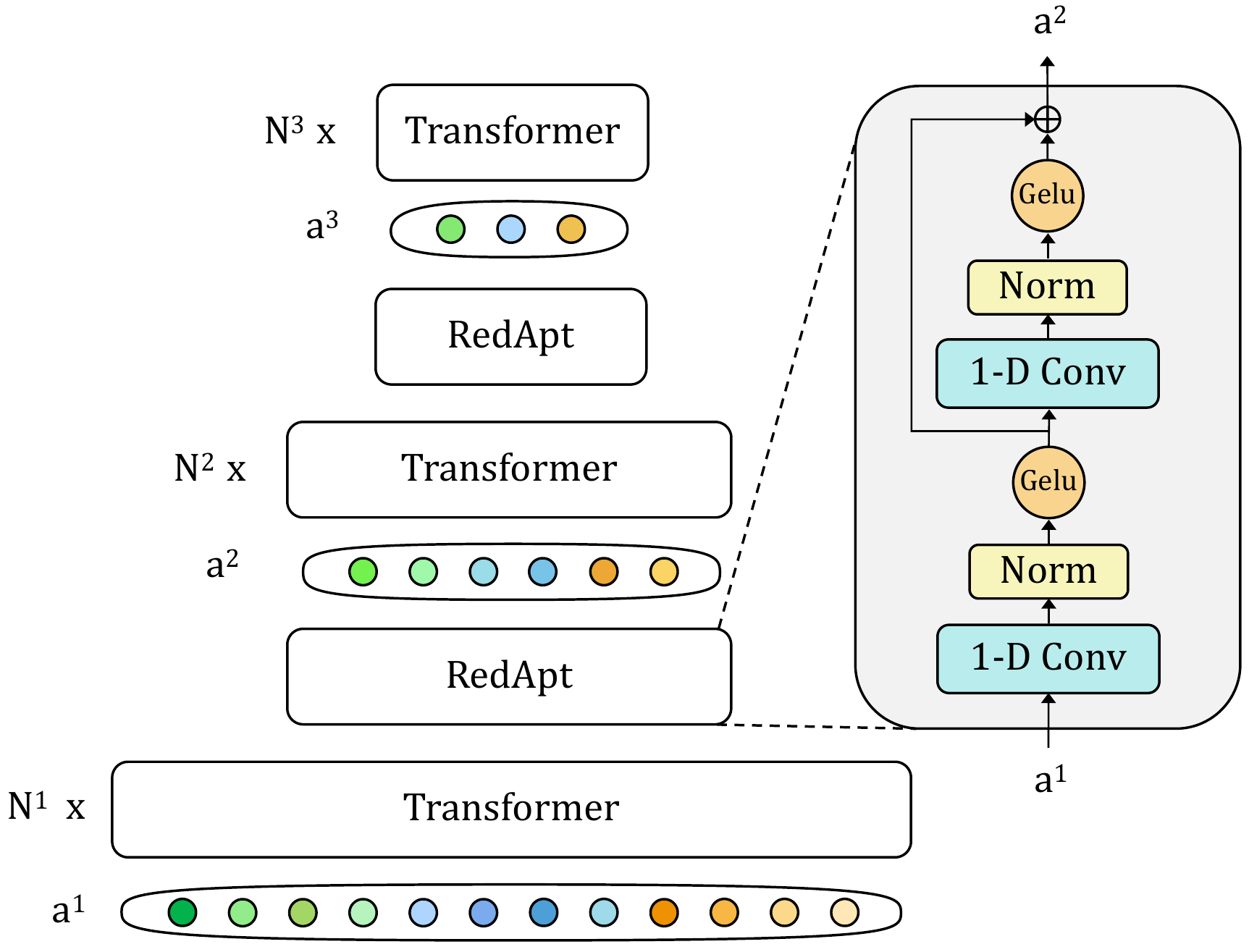}
  \caption{The \model architecture and integration.}
  \label{fig:arch}
\end{figure}

The core idea is to pool a temporal speech sequence, to reduce it length while learning local information from the shrunk sequence. Suppose \textsc{w2v2} feature encoder  yields a sequence, $\textbf{a}$. As shown in Figure~\ref{fig:arch} (\textsc{w2v2} is omitted for brevity), \model is built on two CNN blocks which are wrapped by layer normalization, residual connection, and GELU nonlinear activation. 

The first CNN block is a pooling module to shrink the length of $\textbf{a}$. It is parameterized with kernel $k$, stride length $s$ and padding 
$p$. The input sequence length thus can be reduced,
\begin{equation}\label{reducing}
\setlength{\abovedisplayskip}{3pt}
\setlength{\belowdisplayskip}{3pt}
\setlength{\abovedisplayshortskip}{3pt}
\setlength{\belowdisplayshortskip}{3pt}
    n'=\left\lfloor\frac{n+2 p - k}{s}\right\rfloor+1
\end{equation}
where $n$ and $n'$ are the lengths of the original and shortened sequences, respectively. After GELU activation, we can get  $\textbf{a}'$. The second CNN block learns shared position-wise kernels within a given window which can re-capture local information,
\begin{equation}
\setlength{\abovedisplayskip}{3pt}
\setlength{\belowdisplayskip}{3pt}
\setlength{\abovedisplayshortskip}{3pt}
\setlength{\belowdisplayshortskip}{3pt}
    \mathbf{a}^{\prime \prime}=\mathbf{a}^{\prime}+\operatorname{GELU}\left(\operatorname{Norm}\left(\operatorname{CNN}\left(\mathbf{a}^{\prime}\right)\right)\right)
\end{equation}
The intuition is that certain information (e.g., positional awareness \cite{dai2020funnel}) is lost during pooling, and requires restoration.%
%
%
%
%
The inclusion of layer normalization, residual connection, and  nonlinearity follow the same  rational as Transformer blocks~\cite{NIPS2017_7181}. The total number of parameters for a single block of \model is 11.5M.



\subsection{\model Integration into \textsc{wav2vec 2}}


\paragraph{\textsc{wav2vec 2 (w2v2)}} architecture  starts with a CNN-based feature encoder to  extract features from raw speech signals, while performing downsampling. A quantization module is attached on top of the feature encoder to learn discrete latent speech vectors. The output of the feature encoder is masked and forwarded into a context network, consisting of 24 Transformer blocks and 16 self-attention heads (for the \textsc{large} configuration), to learn contextualized speech representations. The entire model is pretrained with a contrastive loss to distinguish a true masked latent vector from those generated by the model. After pretraining, only the feature extractor and context network are fine-tuned in the downstream tasks. 

\model progressively shrinks a temporal sequence  during the forward propagation in the \textsc{w2v2} Transformer-based encoder. Assuming an integration of $m$ \model blocks into Transformer blocks, 
%
%
%
\model blocks compress a sequence by a factor of $s^m$. As each Transformer block has a quadratic complexity  w.r.t. the input length $n_0$, pooling tensors has significant benefits on memory and compute requirements of \textsc{w2v2}. Denoting the sequence length at i-th layer of \textsc{w2v2} as $n_i$ and the stride size as $s_i$, the complexity of each subsequent Transformer layer (until the next pooling) is $\mathcal{O}((\frac{n_i}{s_i})^2)$ compared to the  $\mathcal{O}(n_0^2)$ in vanilla Transformer block.

\subsection{\model Integration into ST}
We are motivated to capitalize on pre-trained modules, \textsc{w2v2} and mBART \cite{liu2020multilingual}, that were trained on large unlabelled and labelled data. Our ST model consists of the \textsc{w2v2}+\model encoder, and an mBART decoder.~Since the goal of our work is to enable faster signal encoding, we train the encoder while freezing the decoder in all experiments. Unlike~\citet{gallego2021end} that use stage-training, we train \textsc{w2v2} and \model jointly in one step.  

\section{Experiments}
In this section, we first describe our experimental settings (\S\ref{sec_exsetup}). Next, we investigate the effect of the number of \model blocks (\S\ref{sec_number}) and their positions~(\S\ref{sec_position}) on speed, memory, and translation quality. Then, we evaluate our ST model on 8 language pairs from Must-C benchmarks~(\S\ref{sec:translation_results}) and analyze  inference time~(\S\ref{sec_inference}).  Lastly, we provide an  ablation study on \model components~(\S\ref{sec_ablation}).
\subsection{Experimental Settings}\label{sec_exsetup}
\stitle{Dataset.} We use the Must-C dataset \cite{cattoni2021must}, a multilingual ST corpus collected from TED talks. We experimented with 8 language pairs, using English (EN) as source and the followings as target: German (DE), Romanian (RO), Spanish (ES), French (FR), Dutch (NI), Portuguese (PT), Russian (RU), and Italian (IT). The data is preprocessed and filtered following steps outlined in \citet{gallego2021end}. The best systems were selected on dev sets, and results are reported on  test set (tst-COMMON). We use the EN-DE pair for the detailed analysis and ablation. 



\stitle{Implementation Details.}
Similar to the modern ST architectures~\cite{gallego2021end,tang2020multilingual}, we use pretrained \textsc{w2v2} large as our encoder and pretrained mBART50 decoder as the decoder. We randomly initialize the top 3 layers of \textsc{w2v2} in experiments involving \model and find that it enables faster convergence, verifying earlier observations by~\citet{sun2021multilingual}. We free \textsc{w2v2} feature extractor.~For full details on training configurations and hardware, please refer to \emph{Appendix} \ref{appendix:training details}.




\stitle{Baseline.}
We use the SotA ST model from \citet{gallego2021end} as our baseline. The model uses a similar \textsc{w2v2} large encoder, along with a CNN-based length adapter on top of the encoder which reduces the sequence length by a factor of 8. The decoder is similar to ours and is frozen during training for fair comparison between the two models. This model is denoted as \textsc{w2v2+} hereafter. While \model offers various degrees of layer-wise reduction, for comparability in the translation experiments on 8 language pairs (\S\ref{sec:translation_results}), the configuration with the same reduction factor (i.e., 3 blocks of \model) and total encoder parameter size is used.  

\stitle{Metrics.} We measure efficiency at inference in terms of \textit{throughput} (the number of speech data that can be processed in a unit of time), \textit{memory} (GPU memory usage); and \textit{FLOPs} (floating-point operations performed given a single process, higher FLOPs means slower inference speed). 
See \emph{Appendix} \ref{appendix:flops details} for more details. We use BLEU\footnote{\url{https://github.com/mjpost/sacreBLEU}} to evaluate translation quality.


\begin{table}[t]
\small
\centering
  \scalebox{0.9}{
    \begin{tabular}{lccccc}
    \toprule
    $m$  &0 & 1 & 2 & 3 & 4  \\
    \midrule
    BLEU & 26.51 &  27.69 & 27.61 & 27.42  & 25.86  \\
    \midrule
    Throughput $\uparrow$ & 1.00x & 1.22x & 1.26x & 1.31x & 1.35x  \\
    \midrule
    Memory $\downarrow$ & 1.00x & 0.80x & 0.77x & 0.73x & 0.67x     \\
    \midrule
    
    FLOPs $\downarrow$ & 1.00x & 0.86x & 0.84x & 0.81x & 0.76x \\

    \bottomrule
    \end{tabular}}
    \caption{BLEU, throughput, memory consumption, and FLOPs at training and inference for various number of \model blocks: $m=\{1, 2, 3, 4\}$, the positions of blocks are $\{[15], [15,20], [15,18,19], [14,15,18,19]\}$.
    }
    \label{tab:comp}
\end{table}
\subsection{Selecting the Number of \model Blocks}\label{sec_number}
We examine the impact of \model on translation quality and efficiency by injecting various number of \model blocks, $m=\{0, 1, 2, 3, 4\}$, where 0 refers to the baseline and the rest indicate our models. As a heuristic for setting the cap on $m$, we use the maximum representation length reduction that matches that of a corresponding text transcripts to avoid the risk of information loss. The intuition is that given the same content, the length of speech representations (after \textsc{w2v2}) should not be less than that of text representations; the former may vary depending on the degree of compression, whereas the latter is a fixed number. This reflects on the fact that text, unlike speech, carries only the content information that is essential for translation task. Further investigation of text representations optimality compared with speech is beyond our current focus and we leave it to future work.
%
%
%

Table \ref{tab:comp} summarizes the results.\footnote{Different position configurations for each value of $m$ are tried and we report the best results.} Overall, we achieve significant throughput speedup, memory footprint saving and FLOPs reduction as $m$ increases, while the trend follows a law of diminishing returns. All models with $m\leq3$ retained the same level of translation quality as the baseline ($m=0$). Particularly for $m=3$, memory consumption, throughput, and FLOPs are 0.73$\times$, 1.27$\times$, and 0.81$\times$ of the baseline. The gains from reduced computational cost can be re-invested to increase the batch size at inference and we report the changes of these metrics as the batch size varies in \emph{Appendix} \ref{appendix:flops fig}. We observe a trade-off associated with $m$ in translation quality and efficiency, and we set $m$ to 3 in all our following experiments. 
For brevity, we use \model to refer to our ST models in which \model blocks are injected into the encoder, hereafter. In the next section, we will investigate various layers of placing \model blocks.

\begin{table}[!t]
\centering
\small
  \scalebox{0.9}
  {
  \begin{tabular}{lcccccc}
    \toprule
     Model & Positions & BLEU   &  T.put$\uparrow$  & Mem$\downarrow$ & FLOPs$\downarrow$\\
    \midrule
    \addlinespace[0.3em]
     {\small{W2V2+}} &  -  & 26.51  & 1.00x  &  1.00x   &    1.00x \\
     \midrule
\addlinespace[0.3em]
  \multirow{2}{1em}{\model} 
   & [2,5,6]    &  0.7$^\diamond$   & 1.47x &    0.28x &  0.45x\\
   & [7,9,11]        &  22.02   & 1.41x &    0.45x &  0.58x\\
   & [13,15,20]      &  \underline{27.24}   & \underline{1.41x} &    \underline{0.67x} &  \underline{0.76x}\\
   & [14,18,20]      &  26.83   & 1.33x &    0.60x &  0.80x\\
   & [15,18,19]      &  \textbf{27.42}    & 1.31x &    0.73x & 0.81x\\
   & [16,18,20]      & 27.17    & 1.28x &    0.76x &  0.83x\\
   & [17,19,20]      &  26.78   & 1.24x &    0.79x &  0.85x\\

    \bottomrule
  \end{tabular}}
  \caption{Comparison of different position configurations in terms of translation quality, throughput (T.put), memory (Mem) and FLOPs,  at training and inference time. $^\diamond$: models did not converge. \textbf{Bold}: Best BLEU score. \underline{Underline}: Best configuration.
  }
  \label{tab:position} 
\end{table}


\subsection{Selecting the Positions of \model Blocks}\label{sec_position}
We investigated various positions of \model blocks. 
%
%
Since it is not ideal to experiment all $24^m$ choices, to determine optimal positions, we apply a backward selection mechanism starting from the configuration of [14,15,18,19] by removing one position, or replacing it with another position. 
For selection purposes, we segment Transformer networks of \textsc{wav2vec 2} to 2 buckets, i.e., \textit{low-mid} (0-11), \textit{mid-top} (12-23). While we treat positions as hyper-parameter in our work, a more principled approach could frame it as neural architecture search~\cite{elsken2019neural}. We leave further investigation  to future work. 

The majority of the models with exceptional performance come from \textit{mid-high} layers. Table~\ref{tab:position} presents BLEU scores, memory usage, throughput and FLOPs for  EN-DE, with different position configurations. We observe injecting \model into lower level of \textsc{w2v2} leaves a major toll on BLEU. This verifies the earlier findings on text transformers~\cite{goyal2020power} that higher layers typically convey similar overlapping information while disturbing the lower layers could result in a great deal of information loss. Additionally, compressing sequences in lower levels has greater impact on the pre-trained weights in the subsequent layers, which can result in optimization issues.

We choose our best configuration, [13,15,20], which exhibits efficiency improvements of 1.41$\times$ in throughput, 0.67$\times$ in memory usage and 0.76$\times$ with FLOPs, and use it in our next translation experiments on 8 language pairs.

\subsection{Translation Quality on Must-C}\label{sec:translation_results}
Table \ref{tab:other_lang} reports the results for the 8 Must-C language pairs, using our best configuration~(\S\ref{sec_position}). Our ST models outperform the baseline models~(\S\ref{sec_exsetup}) on 5 language pairs by a large margin, while being comparable on the rest. 
On average, we observe a boost of 0.68 BLEU scores across 8 languages. 
We speculate that these gains could be attributed to the positive impact of dimensionality reduction on filtering out redundancy and noise from the representations, verifying the earlier observations by~\citet{zhang2020adaptive}.

\begin{table}[t]
\centering
  \scalebox{0.65}
  {
  \begin{tabular}{lcccccccc}
    \toprule
    Model &DE & RO  & ES & FR & NI & PT & RU &  IT\\
    \midrule
    \textsc{w2v2+} & 26.51 & 24.66 & 30.04 & 36.26 & 31.08 & 32.67  & 17.19 & 22.13\\\hline
    \model  & \textbf{27.24}  & 24.34 &  \textbf{30.49} & \textbf{37.59} &29.82  &  32.65 & \textbf{18.08} & \textbf{25.73}\\
    \bottomrule
  \end{tabular}
  }
  \caption{Translation BLEU of the SotA model (\textsc{w2v2+}) and our model on 8 language pairs from Must-C.}\label{tab:other_lang} 
\end{table}


\subsection{Inference time for ST}\label{sec_inference}
In order to measure the inference time for the entire ST model, we partitioned audio of EN-DE test set into 5 buckets based on length (seconds)
Compared to the baseline, the decoding speedups are 7\%, 7\%, 5\%, 3\%, 3\% for these buckets $(0, 4), [4, 8 ) , [8, 13), [13, 20], (20,\infty)$, which have 1024, 896, 384, 128 and 64 examples, respectively. 
As expected, the efficiency gain in encoding tends to vanish in the full ST setup due to the depth of the mBART decoder (i.e., 12 layers) and the auto-regressive decoding. 

\subsection{Ablation of \model components}\label{sec_ablation}

To study the contribution from each components of \model block beyond the first CNN block~(\S\ref{redapt_arch}), we conduct an ablation by removing the remaining three components one at a time. The ablation (on EN-DE) for our best configuration (\S\ref{sec_position}) indicates that removing the second CNN block leads to 0.44 BLEU decay, while removing either the GELU and LayerNorm leads to convergence issues (neither models converge). We report further details and positions in \emph{Appendix} \ref{appendix:ablation}.

\section{Conclusion}
\label{sec:conclusion}

We proposed a novel dimensionality reduction block, \model, to improve the efficiency of pretrained speech encoders, e.g. \textsc{wav2vec 2 (w2V2)}, in speech translation (ST). We demonstrated that the integration of \model brings $1.41\times$, $0.67\times$ and $0.76\times$ in speedup, memory usage and  FLOPs reduction at inference. Meanwhile, compared with SotA, our ST system on average improves translation quality by 0.68 BLEU scores over 8 Must-C language pairs. As our future work, we will be investigating the impact of \model in other speech processing tasks~\cite{DBLP:conf/interspeech/YangCCLLLLSCLHT21}, as well as learning the optimal positions for injecting \model blocks via neural architecture search.


\section{Limitations}
While hardware requirement is a common challenge shared across all modern ST models, it is worth mentioning that our work requires GPUs with 16 GB of memory for inference and 48 GB memory for training.




\section{Ethics Statement}
Our work is leveraging pretrained models of language (\textsc{wav2vec2} for speech, mBART for text). However, our method is not designed or intended to rectify any of the well-documented  issues of such models. Hence, our work inherits similar potential risks that these models pose.

\bibliography{custom}
\bibliographystyle{acl_natbib}

\clearpage
\appendix
\section{Appendix}
\label{sec:appendix}
\subsection{Dataset}\label{appendix:data details}
We use all 8 language pairs from Must-C. Please refer to \citet{cattoni2021must} for details for the size of the datasets and train/dev/test splits. We adopt the filtering techniques proposed in \citet{gallego2021end}. We remove instances whose audios are over 25 seconds. We then filter out examples whose transcriptions that 1) have speaker names and non-textual events; 2) start with certain patterns indicating noise. Next, we apply ASR to audios and remove those whose ASR outputs have low WER scores. 

To diversify training data, we also apply data augmentation on the audio data on-the-fly, an effective technique used in ST \cite{potapczyk2019samsung}. We apply "tempo" and "pitch" to make the model become invariant to speaking speeed, and "echo" to simulate echoing. Each instance is augmented with a probability of 0.8 where all effects are applied. We then normalize it to zero mean and unit variance. The parameters of tempo, pitch, echo-delay and echo-decay are to set to (0.85, 1.3), (-300, 300), (20, 200) and (0.05, 0.2).

\subsection{Training Details}\label{appendix:training details}
All models are trained with fairseq \cite{ott2019fairseq} on 4 RTX 6000 GPUs, using 16 floating point precision, for $25k$ updates. We use \textsc{Wav2vec 2}\footnote{\url{https://dl.fbaipublicfiles.com/fairseq/wav2vec/wav2vec\_vox\_960h\_pl.pt}} and the mBart50\footnote{\url{https://dl.fbaipublicfiles.com/fairseq/models/mbart50/mbart50.ft.1n.tar.gz}}  decoder. We limit the source and target lengths and to 400k (i.e., 25 seconds) and 1,024 tokens, respectively.  We use an Adam optimizer with \cite{DBLP:journals/corr/KingmaB14} $\beta_1 = 0.99$ and  $\beta_2 = 0.98$. We set the dropout to 0.1, clip norm to 20, and the label smoothing value to 0.2. For the baseline models, we use a learning rate of 5e-04 and reduce it when loss stops improving. Depending on speech lengths for each source language, we set the average batch size being either 64, or 128. For our models, we use a learning rate of 5e-04 for DE and NL, 4e-04 for Fr and 3e-04 for the rest, and we also decrease the learning rate at plateau. We use an effective batch size of 64 for all language pairs. We set kernel size, stride and padding for the two CNN blocks in \model to <3, 2, 1> and <3, 1, 1>.  We report BLEU results on single models without checkpoint averaging.

\begin{figure}[!t]
    \centering
    \includegraphics[width=0.5\textwidth]{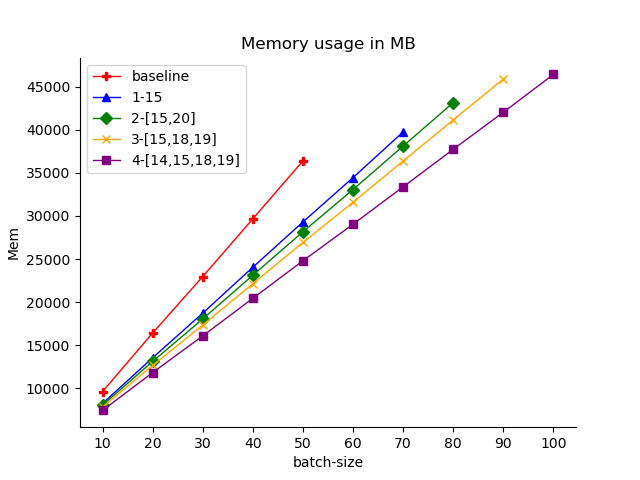}
    \includegraphics[width=0.5\textwidth]{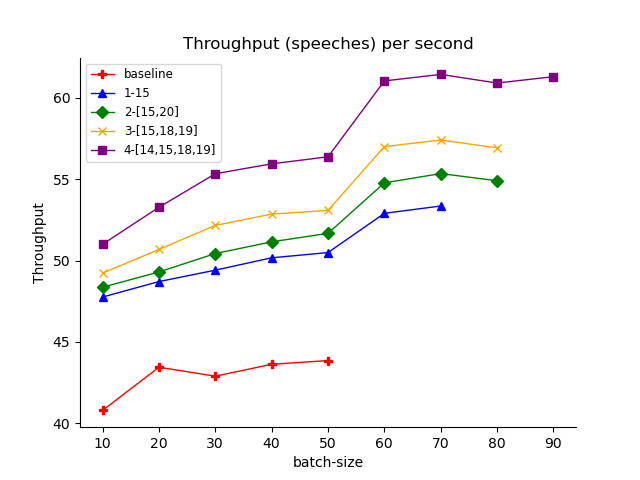}
    \includegraphics[width=0.5\textwidth]{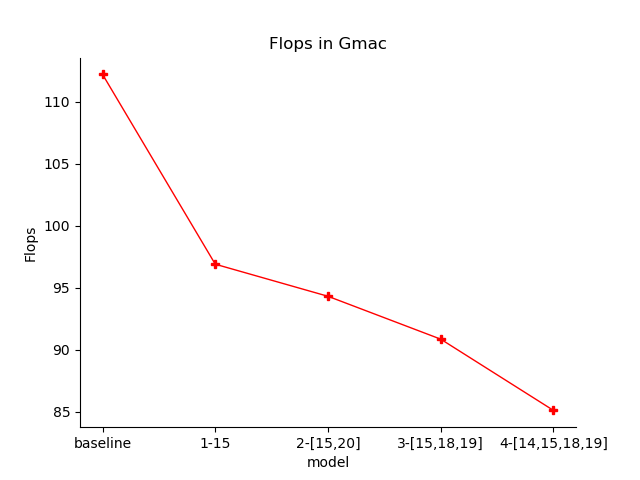}
    \caption{ memory usage (Upper)  throughput (middle) and  FLOPs (bottom)}
    \label{fig:3plot}
\end{figure}

\subsection{Memory and Throughput vs. Batch Size}\label{appendix:flops fig}
See Figure \ref{fig:3plot} (upper, middle) for memory usage and throughput (bottom) for each model when we increase batch size until GPU memory gets full. Our methods have better utilization of GPU memory.

\subsection{Measuring Throughput and FLOPs}\label{appendix:flops details}
We set the length of raw signals as 88,000, which is the average signals length in the training set, and the batch size as 64 when computing throughput and memory. 
We use batch size 1 for calculating FLOPs, whose value is mainly affected by the input length. All testings are performed on one RTX 8000 GPU.

\subsection{Ablation of \model components}\label{appendix:ablation}

Table \ref{tab:abs} shows the effect of removing each components in \model.

\begin{table}[!htbp]
\centering
\small
  \scalebox{0.8}{
    \begin{tabular}{lcccc}
    \toprule
    \# Positions & 2nd CNN & LayerNorm & GELU & BLEU    \\
    \midrule
    13-15-20    & \checkmark & \checkmark  &  \checkmark & 27.24 \\
                & - &  \checkmark &  \checkmark & 26.80 \\
                & \checkmark & -  &  \checkmark & $^\diamond$  \\
                & \checkmark & -  &  -  & $^\diamond$ \\
    \midrule
    15-18-19    & \checkmark & \checkmark  &  \checkmark & 27.42 \\
                & - &  \checkmark & \checkmark & 27.11 \\
                & \checkmark & -  &  \checkmark & $^\diamond$  \\
                & \checkmark & -  &  -  & $^\diamond$ \\        
    \bottomrule
    \end{tabular}}
    \caption{Ablation on Acoustic Pooler. $^\diamond$: models did not converge.
    }
    \label{tab:abs}
\end{table}


\end{document}